\documentclass[11pt,a4paper]{article}
\usepackage[hyperref]{acl2019}
\usepackage{times}
\usepackage{latexsym}

\usepackage{url}

\usepackage{amsthm}
\usepackage{amsmath}
\usepackage{subfiles}
\usepackage{graphicx}
\usepackage{pgfplots}
\pgfplotsset{width=7cm,compat=1.9}
\usepackage{xcolor}
\definecolor{mildgreen1}{RGB}{232, 255, 232}
\definecolor{mildgreen2}{RGB}{242, 255, 242}
\definecolor{faintgreen}{RGB}{247, 255, 247}
\definecolor{hardgreen}{RGB}{204, 255, 204}
\definecolor{mildred1}{RGB}{25, 0, 0}
\definecolor{mildred2}{RGB}{21, 0, 0}
\definecolor{faintred1}{RGB}{255, 250, 250}
\definecolor{faintred2}{RGB}{255, 242, 242}
\definecolor{faintred3}{RGB}{255, 230, 230}
\definecolor{faintred4}{RGB}{255, 220, 220}
\definecolor{hardred}{RGB}{255, 190, 190}

\definecolor{faintred5}{RGB}{255, 80, 80}

\usepackage{multirow}

\theoremstyle{definition}
\newtheorem{definition}{Definition}[section]

\aclfinalcopy 
\def\aclpaperid{2203} 

\newcommand\integratedgrads{\textit{IG}}
\newcommand{\synteq}{::=}

\title{Incorporating Priors with Feature Attribution on Text Classification}

\author{Frederick Liu \qquad Besim Avci \\
  Google  \\
  \texttt{\{frederickliu, besim\}@google.com}}

\date{}

\begin{document}
\maketitle
\begin{abstract}
  Feature attribution methods, proposed recently, help users interpret the predictions of complex models. 
  Our approach integrates feature attributions into the objective function to allow machine learning practitioners to incorporate priors in model building.
  To demonstrate the effectiveness our technique, we apply it to two tasks: (1) mitigating unintended bias in text classifiers by neutralizing identity terms; (2) improving classifier performance in a scarce data setting by forcing the model to focus on toxic terms. Our approach adds an $L_2$ distance loss between feature attributions and task-specific prior values to the objective. 
  Our experiments show that $i$) a classifier trained with our technique reduces undesired model biases without a tradeoff on the original task; $ii$) incorporating priors helps model performance in scarce data settings. 
\end{abstract}

\section{Introduction}

One of the recent challenges in machine learning (ML) is interpreting the predictions made by models, especially deep neural networks. Understanding models is not only beneficial, but necessary for wide-spread adoption of more complex (and potentially more accurate) ML models. 
From healthcare to financial domains, regulatory agencies mandate entities to provide explanations for their decisions \cite{Goodman2017EuropeanUR}. Hence, most machine learning progress made in those areas is hindered by a lack of model explainability -- causing practitioners to resort to simpler, potentially low-performance models. To supply for this demand, there has been many attempts for model interpretation in recent years for tree-based algorithms \cite{LundbergTree} and deep learning algorithms \cite{lundberg2017unified, SmilkovTKVW17,pmlr-v70-sundararajan17a, Bach2015OnPE, KimWGCWVS18, DhurandharCLTTS18}. On the other hand, the amount of research focusing on explainable natural language processing (NLP) models \cite{Jiwei, Murdoch, lei2016rationalizing} is modest as opposed to image explanation techniques.

\begin{table}[!t]
    \begin{tabular}{ c | l | c }
        \textbf{Method} &  \multicolumn{1}{c|}{\textbf{Sentence}} & \textbf{Probability} \\ \hline 
        \multirow{2}{*}{Baseline} & \colorbox{faintred1}{I} \colorbox{faintred1}{am} \textit{\colorbox{faintred5}{gay}} & 0.915 \\
         & \colorbox{faintred2}{I} \colorbox{faintred3}{am} \colorbox{hardred}{straight} & 0.085 \\ \hline 
        \multirow{2}{*}{Our Method} & \textit{\colorbox{faintred2}{I}} \textit{\colorbox{faintred1}{am}} \colorbox{faintred1}{gay} & 0.141 \\
         & \textit{\colorbox{faintred1}{I}} \textit{\colorbox{faintred2}{am}} \colorbox{faintred1}{straight} & 0.144
    \end{tabular}
    \centering
    \caption{Toxicity probabilities for samples of a baseline CNN model and our proposed method. Words are shaded based on their attribution and italicized if attribution is $>0$.}
    \label{table:motivation}
\end{table}

Inherent problems in data emerge in a trained model in several ways. Model explanations can show that the model is not inline with human judgment or domain expertise. A canonical example is model unfairness, which stems from biases in the training data.
Fairness in ML models rightfully came under heavy scrutiny in recent years \cite{ZhangBias, DixonLSTV18, Angwin}.
Some examples include sentiment analysis models weighing negatively for inputs containing identity terms such as ``jew'' and ``black'', and hate speech classifiers leaning to predict \emph{any} sentence containing ``islam'' as toxic \cite{Waseem-2013}. 
If employed, explanation techniques help divulge these issues, but fail to offer a remedy. For instance, the sentence ``I am gay'' receives a high score on a toxicity model as seen in Table \ref{table:motivation}. The Integrated Gradients \cite{pmlr-v70-sundararajan17a} explanation method attributes the majority of this decision to the word ``gay.'' However, none of the explanations methods suggest next steps to fix the issue. Instead, researchers try to reduce biases indirectly by mostly adding more data \cite{DixonLSTV18, ChenJS18}, using unbiased word vectors \cite{Park2018ReducingGB}, or directly optimizing for a fairness proxy with adversarial training \cite{MadrasCPZ18, ZhangBias}. These methods either offer to collect more data, which is costly in many cases, or make a tradeoff between original task performance and fairness.

In this paper, we attempt to enable injecting priors through model explanations to rectify issues in trained models. We demonstrate our approach on two problems in text classification settings: (1) model biases towards protected identity groups; (2) low classification performance due to lack of data. The core idea is to add $L_2$ distance between Path Integrated Gradients attributions for pre-selected tokens and a target attribution value in the objective function as a loss term. For model fairness, we impose the loss on keywords identifying protected groups with target attribution of $0$, so the trained model is penalized for attributing model decisions to those keywords. Our main intuition is that undesirable correlations between toxicity labels and instances of identity terms cause the model to learn unfair biases which can be corrected by incorporating priors on these identity terms. 
Moreover, our approach allows practitioners to impose priors in the other direction to tackle the problem of training a classifier when there is only a small amount of data. As shown in our experiments, by setting a positive target attribution for known toxic words \footnote{Full list of identity terms and toxic terms used as priors can be found in supplemental material. Please note the toxic terms are not censored.}, one can improve the performance of a toxicity classifier in a scarce data regime.

We validate our approach on the Wikipedia toxic comments dataset \cite{Wulczyn:2017}. Our fairness experiments show that the classifiers trained with our method achieve the same performance, if not better, on the original task, while improving AUC and fairness metrics on a synthetic, unbiased dataset. Models trained with our technique also show lower attributions to identity terms on average. Our technique produces much better word vectors as a by-product when compared to the baseline. Lastly, by setting an attribution target of $1$ on toxic  words, a classifier trained with our objective function achieves better performance when only a subset of the data is present. 

\section{Feature Attribution}
In this section, we give formal definitions of feature attribution and a primer on [Path] Integrated Gradients (IG), which is the basis for our method.

\begin{definition}{}\label{def:def1}
Given a function $f:R^n \rightarrow [0, 1]$ that represents a model, and an input $\boldsymbol{x} = (x_1,..., x_n) \in R^n$. An attribution of the prediction at input $\boldsymbol{x}$ is a vector $\boldsymbol{a} = (a_1,..., a_n)$ and $a_i$ is defined as the attribution of $x_i$.
\end{definition}

Feature attribution methods have been studied to understand the contribution of each input feature to the output prediction score. This contribution, then, can further be used to interpret model decisions. Linear models are considered to be more desirable because of their implicit interpretability, where feature attribution is the product of the feature value and the coefficient. To some, non-linear models such as gradient boosting trees and neural networks are less favorable due to the fact that they do not enjoy such transparent contribution of each feature and are harder to interpret \cite{Lou:2012}.   

Despite the complexity of these models, prior work has been able to extract attributions with gradient based methods \cite{SmilkovTKVW17}, Shapley values from game theory (SHAP) \cite{lundberg2017unified}, or other similar methods \cite{Bach2015OnPE, shrikumar17a}. Some of these attributions methods, for example Path Intergrated Gradients and SHAP, not only follow Definition ~\ref{def:def1}, but also satisfy axioms or properties that resemble linear models. One of these axioms is completeness, which postulates that the sum of attributions should be equal to the difference between uncertainty and model output. 

\subsection*{Integrated Gradients}

\textit{Integrated Gradients} \cite{pmlr-v70-sundararajan17a} is a model attribution technique applicable to all models that have differentiable inputs w.r.t. outputs. IG produces feature attributions relative to an uninformative baseline. This baseline input is designed to produce a high-entropy prediction representing uncertainty. IG, then, interpolates the baseline towards the actual input, with the prediction moving from uncertainty to certainty in the process. Building on the notion that the gradient of a function, $f$, with respect to input can characterize sensitivity of $f$ for each input dimension, IG simply aggregates the gradients of $f$ with respect to the input along this path using a path integral. The crux of using path integral rather than overall gradient at the input is that $f$'s gradients might have been saturated around the input and integrating over a path alleviates this phenomenon. Even though there can be infinitely many paths from a baseline to input point, \textit{Integrated Gradients} takes the straight path between the two. We give the formal definition from the original paper in \ref{def:def2}.
\theoremstyle{definition}
\begin{definition}{}\label{def:def2}
Given an input $\boldsymbol{x}$ and baseline
$\boldsymbol{x'}$, the integrated gradient along the $i^{th}$ dimension is defined as follows.
\begin{equation}\label{eqn:intgrad}
\begin{aligned}
& \integratedgrads_{i}(\boldsymbol{x},\boldsymbol{x'}) \synteq \\
& (x_i-{x'}_i)\times\int_{\alpha=0}^{1} \tfrac{\partial f(\boldsymbol{x'} + \alpha\times(\boldsymbol{x}-\boldsymbol{x'}))}{\partial x_i  }~d\alpha
\end{aligned}
\end{equation}
where
$\tfrac{\partial f(\boldsymbol{x})}{\partial x_i}$ represents the gradient of
$f$ along the $i^{th}$ dimension at $\boldsymbol{x}$.
\end{definition}
In the NLP setting, $\boldsymbol{x}$ is the concatenated embedding of the input sequence. The attribution of each token is the sum of the attributions of its embedding.

There are other explainability methods that attribute a model's decision to its features, but we chose IG  in this framework due to several of its characteristics. First, it is both theoretically justified  \cite{pmlr-v70-sundararajan17a} and proven to be effective in NLP-related tasks \cite{PramodVQA}. Second, the IG formula in \ref{def:def2} is differentiable everywhere with respect to model parameters. Lastly, it is lightweight in terms of implementation and execution complexity.

\section{Incorporating Priors}
Problems in data manifest themselves in a trained model's performance on classification or fairness metrics. Traditionally, model deficiencies were addressed by providing priors through extensive feature engineering and collecting more data. Recently, attributions help uncover deficiencies causing models to perform poorly, but do not offer actionability.

To this end, we propose to add an extra term to the objective function to penalize the $L_2$ distance between model attributions on certain features and target attribution values. This modification allows model practitioners to inject priors. For example, consider a model that tends to predict every sentence containing ``gay'' as toxic in a comment moderation system. Penalizing non-zero attributions on the tokens identifying protected groups would force the model to focus more on the context words rather than mere existence of certain tokens. 

We give the formal definition of the new objective function that incorporates priors as the follows:
\begin{definition}{}\label{def:def3}
Given a vector $\boldsymbol{t}$ of size $n$, where $n$ is the length of the input sequence and $t_{i}$ is the attribution target value for the $i$th token in the input sequence. The \emph{prior} loss for a scalar output is defined as:
\begin{equation}\label{eqn:prior}
\begin{aligned}
\mathcal{L}^{\emph{prior}}(\boldsymbol{a}, \boldsymbol{t}) = 
\sum_{i}^{\textit{n}}{(a_{i} - t_{i})^2} 
\end{aligned}
\end{equation}
where $a_{i}$ refers to attribution of the $i$th token as in Definition \ref{def:def1}. 
\end{definition}
For a multi-class problem, we train our model with the following joint objective,

\begin{equation}\label{eqn:joint}
\begin{aligned}
\mathcal{L}^{\emph{joint}} = \mathcal{L}(\boldsymbol{y},\boldsymbol{p}) + \lambda   \sum_c^C \mathcal{L}^{\emph{prior}}(\boldsymbol{a^c}, \boldsymbol{t^c})
\end{aligned}
\end{equation}
where $\boldsymbol{a^c}$ and $\boldsymbol{t^c}$ are the attribution and attribution target for class $c$, $\lambda$ is the hyperparameter that controls the stength of the prior loss and $\mathcal{L}$ is the cross-entropy loss defined as follows:
\begin{equation}\label{eqn:cross}
\begin{aligned}
\mathcal{L}(\boldsymbol{y},\boldsymbol{p}) = \sum_{c}^C - y_c \log(p_c)
\end{aligned}
\end{equation}
where $\boldsymbol{y}$ is an indicator vector of the ground truth label and $p_c$ is the posterior probability of class $c$.

The \emph{joint} objective function is differentiable w.r.t. model parameters when attribution is calculated through Equation~\ref{eqn:intgrad} and  can be trained with most off-the-shelf optimizers. 
The proposed objective is not dataset-dependent and is applicable to different problem settings such as sentiment classification, abuse detection, etc. It only requires users to specify the target attribution value for tokens of interest in the corpus. We illustrate the effectiveness of our method by applying it to a toxic comment classification problem. In the next section, we first show how we set the target attribution value for identity terms to remove unintended biases while retaining the same performance on the original task. Then, using the same technique, we show how to set target attribution for toxic words to improve classifier performance in a scarce data setting.

\section{Experiments}
\label{exp}
\begin{table}
    \begin{tabular}{c|ccccc}
    \textbf{Identity} & \textbf{Base}  & \textbf{Imp} & \textbf{TOK} & \textbf{Ours}  \\ \hline
    gay            & .272 &  .353  & -.006 & .000               \\ \hline
    homosexual            & .085 & .388 & -.006 & -.000                \\ \hline
    queer            & .071 & .28 & -.006 & .000             \\ \hline
    teenage            & .030     & -0.02 & -.006 & -.001           \\ \hline
    lesbian            & .012     & .046 & -.006 & .001 \\ \hline\hline
    vocab avg  & -.002 & -.001 & -.004 & -.001
    \end{tabular}
    \caption{Subset of identity terms we used and their mean attribution value on the test set. Method names are abbreviated with the prefix. The last row is the average across all vocabularies.}
    \label{table:identity}
\end{table}

We incorporate human prior in model building on two applications.  
First, we tackle the problem of unintended bias in toxic comment classification \cite{DixonLSTV18} with our proposed method. 
For our experiments, we aim to mitigate the issue of neutral sentences with identity terms being classified as toxic for a given a set of identity terms. A subset of the identity terms are listed in the first column of Table \ref{table:identity}. Second, we force the model to focus on a list of human-selected toxic terms under scarce data scenario to increase model performance.  

In the following section, we introduce the dataset we train and evaluate on along with a synthetic dataset to further validate our fairness improvements. After that, we describe our experimental setup. Then, we compare our method to a classifier trained without the prior loss and 2 other baselines. Lastly, we show the results demonstrating usefulness of our approach with data scarcity.

\subsection{Dataset}
In this work, we use a dataset containing comments from Wikipedia Talk Pages \cite{DixonLSTV18}. Number of samples are 95,692 / 32,128 / 31,866 in the train/dev/test sets respectively. The ratio of positive (toxic) labels in the training set is 9.7\%. The dataset was annotated by human raters, where toxicity was defined as a ``rude, disrespectful, or unreasonable comment that is likely to make you leave a discussion'' per \newcite{DixonLSTV18}. Please refer to the corresponding paper for more details about collection methodology, biases present in the data, and toxicity distribution per comment length. 

\begin{table}
    \centering
    \begin{tabular}{l|c}
        \textbf{Template Examples} & \textbf{Labels}    \\ \hline
        I am \textbf{Identity} & Non-toxic \\ \hline
        Hug \textbf{Identity} & Non-toxic \\ \hline
        Being \textbf{Identity} is wonderful & Non-toxic \\ \hline
        I hate \textbf{Identity}  & Toxic     \\ \hline
        You are a moronic \textbf{Identity} & Toxic \\ \hline
        \textbf{Name} is a disgusting \textbf{Identity} & Toxic    
    \end{tabular}
    \caption{Example templates used to generate the synthetic dataset for evaluation.}
    \label{table:template}
\end{table}

We also use a synthetically generated dataset to validate our approach on fairness as in \newcite{Park2018ReducingGB, DixonLSTV18}. The dataset is created using a set of templates, developed by \newcite{DixonLSTV18} \footnote{\url{https://git.io/fjVEo}}, where placeholders are replaced with different identity terms. We show a subset of example templates in Table \ref{table:template} and selected identity terms along with their mean attributions across the test set in Table \ref{table:identity}. We mainly evaluate the effectiveness of our debiasing technique on this dataset because the original test sets follow the same biased distribution. Intuition is that predictions returned for sentences containing different identity terms in the exact same context should be similar. Hence, this dataset enables us to quantify the performance of a classifier in more detail when controlled on identity.

\subsection{Experimental Setup}
For the text classifier, we built a convolutional neural network (CNN) classifier as in \newcite{Kim14f}. The network contains a convolution layer with 128 2-, 3-, 4-gram filters for a sequence length of 100 followed by a max-pooling layer and softmax function. Embeddings were randomly initialized and their size was set to 128. Shorter sequences are padded with \textless{}pad\textgreater{} token and longer sequences are truncated. Tokens occurring 5 times or more are retained in the vocabulary. We set dropout as 0.2 and used Adam \cite{KingmaB14} as our optimizer with initial learning rate set to $0.001$. We didn't perform extensive network architecture search to improve the performance as it is a reasonably strong classifier with the initial performance of 95.5\% accuracy.

\begin{table}
  \resizebox{0.48\textwidth}{!}{%
  \begin{tabular}{c | c c c c c c c} 
        \textbf{Whole Dataset} \hspace{0.05cm} & \textbf{Acc} \hspace{0.05cm} & \textbf{F1} \hspace{0.05cm} & \textbf{AUC}  & \textbf{FP} & \textbf{FN} \\ \hline
     Baseline \hspace{0.05cm} &   .955  \hspace{0.05cm} & .728 \hspace{0.05cm} & .948 & .010 & .035\\ 
     Importance \hspace{0.05cm} &   .957 \hspace{0.05cm} & .739 \hspace{0.05cm} & .953  & .009 & .034  \\ 
     TOK Replace\hspace{0.05cm} &   .939 \hspace{0.05cm} & .607 \hspace{0.05cm} & .904  & .014 & .047  \\ 
     \hline
     Our Method \hspace{0.05cm} &   \textbf{.958} \hspace{0.05cm} & \textbf{.752} \hspace{0.05cm} & \textbf{.960} & .009 & \textbf{.032}  \\
    Fine-tuned \hspace{0.05cm} &   .955 \hspace{0.05cm} & .720 \hspace{0.05cm} & .954  & \textbf{.007} & .038 \\
    
  \end{tabular}}
    \caption{Performance on the Wikipedia toxic comment dataset. Columns represent Accuracy, F-1 score, Area Under ROC curve, False Positive, and False Negative. Numbers represent the mean of 5 runs. Maximum variance is $.012$.} 
    \label{table:all}
\end{table}

The number of interpolating steps for IG is set to $50$ (as in the original paper) for calculating Riemann approximation of the integral. Since the output of the binary classification can be reduced to a single scalar output by taking the posterior of the positive (toxic) class, the prior is only added to the positive class in equation \ref{eqn:joint} . We set
\begin{equation}
t_{i}= 
\begin{cases}
    k,& \text{if } x_{i} \in I\\
    a_{i},              & \text{otherwise}
\end{cases},
\end{equation}
where $I$ is the set of selected terms and $x_{i}$ being the $i$ th token in the sequence.

For fairness experiments, we set $k$ to be $0$ and $I$ to the set of identity terms with the hope that these terms should be as neutral as possible when making predictions. Hyperparamter $\lambda$ is searched in the range of $(1, 10^8)$ and increased from 1 by a scale of 10 on the dev set and we pick the one with best F-1 score. $\lambda$ is set to $10^6$ for the final model. 

For data scarcity experiments, we set $k$ to $1$ and $I$ to the set of toxic terms to force the model to make high attributions to these terms. Hyperparameter $\lambda$ is set to $10^5$ across all data size experiments by tuning on the dev set with model given $1\%$ of training data. 

Each experiment was repeated for 5 runs with 10 epochs and the best model is selected according to the dev set.  Training takes 1 minute for a model with cross-entropy loss and 30 minutes for a model with \emph{joint} loss on an NVidia V100 GPU. However, reducing the step size in IG for calculating Riemann approximation of the integral to 10 steps reduces the training time to 6 minutes. Lastly, training with \emph{joint} loss reaches its best performance in later epochs than training with cross-entropy loss.

\begin{table}
    \resizebox{0.48\textwidth}{!}{%
    \begin{tabular}{c | c c c c c}  
        \textbf{Identity} \hspace{0.05cm} & \textbf{Acc} \hspace{0.05cm} & \textbf{F1} \hspace{0.05cm} & \textbf{AUC} & \textbf{FP} & \textbf{FN} \\ \hline
        Baseline \hspace{0.05cm} &   .931 \hspace{0.05cm} & .692 \hspace{0.05cm} & .910 & .011 & .057\\ 
         Importance \hspace{0.05cm} &   .933 \hspace{0.05cm} & \textbf{.704} \hspace{0.05cm}& .945  & .012 & \textbf{.055}  \\ 
         TOK Replace\hspace{0.05cm} &   .910 \hspace{0.05cm} & .528 \hspace{0.05cm} & .882 &  .008 & .081  \\ 
         \hline
         Our Method \hspace{0.05cm} &   \textbf{.934} \hspace{0.05cm} & .697 \hspace{0.05cm}  & \textbf{.949} & .008 & .058\\
        Finetuned \hspace{0.05cm} &   .928 \hspace{0.05cm} & .660 \hspace{0.05cm}  & .940  & \textbf{.007} & .064\\
    \end{tabular}}
    \caption{Performance statistics of all approaches on the Wikipedia dataset filtered on samples including identity terms. Numbers represent the mean of 5 runs. Maximum variance is $.001$.}
    \label{table:filtered}
\end{table}

\subsubsection*{Implementation Decisions}
When taking the derivative with respect to the loss, we treat the interpolated embeddings as constants. Thus, the \emph{prior} loss does not back-propagate to the embedding parameters. There are two reasons that lead to this decision: ($i$) taking the gradient of the interpolate operation would break the axioms that IG guarantees; ($ii$) the Hessian of the embedding matrix is slow to compute. The implementation decision does not imply that \emph{prior} loss has no effect on the word embeddings, though. During training, the model parameters are updated with respect to both losses. Therefore, the word embeddings had to adjust accordingly to the new model parameters by updating the embedding parameters with cross-entropy loss.

\subsection{Results on Incorporating Fairness Priors}
We compare our work to 3 models with the same CNN architecture, but different training settings:
\begin{itemize}
    \item \textbf{Baseline:} A baseline classifier trained with cross-entropy loss.
    \item \textbf{Importance:} Classifier trained with cross-entropy loss, but the loss for samples containing identity words are weighted in the range $(1, 10^8)$, where the actual coefficient is determined to be $10$ on the dev set based on F-1 score.
    \item \textbf{TOK Replace:} Common technique for making models blind to identity terms \cite{Garg2018CounterfactualFI}. All identity terms are replaced with a special \textbf{\textless{}id\textgreater{}} token.
\end{itemize}

We also explore a different training schedule for cases where a model has been trained to optimize for a classification loss:
\begin{itemize}
    \item \textbf{Finetuned:} An already-trained classifier is finetuned with joint loss for several epochs. The aim of this experiment is to show that our method is also applicable for tweaking trained models, which could be useful if the original had been trained for a long time.
\end{itemize}

\begin{table*}[!t]
\centering
\resizebox{0.9\textwidth}{!}{%
    \begin{tabular}{c|c|c||c|c|c||c}
        \multicolumn{3}{c||}{\textbf{gay}} & \multicolumn{3}{c||}{\textbf{homosexual}} & \textbf{\textless{}id\textgreater{}} \\ \hline
        Baseline & Our method & Importance & Baseline & Our method & Importance & Tok Replace \\ \hline \hline
        a**hole            & \textless{}pad\textgreater{}& sh*t & b*tch & scorecard & f*ck & 456\\ \hline
        f*ck            & jus & f*cking & cr*p & dutchman & b*tch & messengers\\ \hline
        pathetic            & tweaking & b*tch & f*g  & `oh  & pu**y & louie \\ \hline
        fu*king            & sess &  f*ck & bulls*** & 678 & sucks & dome\\ \hline
        fa**ot            & ridiculous & penis & dumba*s & nitrites & f*cked & accumulation\\ \hline
        bas**rd            & `do & suck & sh*t & poured & pathetic & ink\\ \hline
        cr*p            & manhood & pu**y & penis & nuts & c*ck & usher\\ \hline
        suck            & dub & d*ckhead & moron & gubernatorial & fart & wikiepedia\\ \hline
        sh*t            & heartening & moron & retard & convincing & a**hole & schizophrenics\\ \hline
        a*s            & desire & fa**ot & gay & strung & fa**ot & notables\\
    \end{tabular}}
    \caption{Top 10 nearest neighbors for tokens `gay' and `homosexual' and \textless{}id\textgreater{} for TOK Replace. All asterisks are inserted by authors to replace certain characters.}
    \label{table:nn}
\end{table*}

\subsubsection{Evaluation on Original Data}

\begin{table}
  \begin{tabular}{c | c c c} 
        \textbf{Synthetic}  & \textbf{AUC}  & \textbf{FPED}  & \textbf{FNED} \\ \hline
     Baseline  &   .885   & 2.77  & 3.51    \\ 
     Importance  &   .850  & 2.90  & 3.06        \\ 
     TOK Replace &   .930 & \textbf{0.00}  &  \textbf{0.00}       \\ 
     \hline
     Our Method  &   \textbf{.952}  & 0.01  & 0.11   \\
    Finetuned  &   .925  & 0.00  & 0.19  \\
    
  \end{tabular}
    \caption{AUC and Bias mitigation metrics on synthetic dataset. The lower the better for Bias mitigation metrics and is bounded by 0. Numbers represent the mean of 5 runs. Maximum variance is $0.013$.}
    \label{table:synth}
\end{table}

We first verify that the prior loss term does not adversely affect overall classifier performance  on the main task using general performance metrics such as accuracy and F-1. Results are shown in Table \ref{table:all}. Unlike previous approaches \cite{Park2018ReducingGB, DixonLSTV18, MadrasCPZ18}, our method does not degrade classifier performance (it even improves) in terms of all reported metrics. We also look at samples containing identity terms. Table \ref{table:filtered} shows classifier performance metrics for such samples. 

The importance weighting approach slightly outperforms the baseline classifier. Replacing identity words with a special tokens, on the other hand, hurts the performance on the main task. One of the reasons might be that replacing all identity terms with a token potentially removes other useful information model can rely on. If we were to make an analogy between the token replacement method and hard ablation, then the same analogy can be made between our method and soft ablation. Hence, the information pertaining to identity terms is not completely lost for our method, but come at a cost.

Results for fine-tuning experiments show the performance after 2 epochs. It is seen that the model converges to similar performance with joint training after only 2 epochs, albeit being slightly poorer. 

\subsubsection{Evaluation on Synthetic Data}
Now we run our experiments on the template-based synthetic data. As stated, this dataset is used to measure biases in the model since it is unbiased towards identities. We use AUC along with False Positive Equality Difference (FPED) and False Negative Equality Difference (FNED), which measure a proxy of Equality of Odds \cite{Hardt:2016}, as in \newcite{DixonLSTV18, Park2018ReducingGB}. FPED sums absolute differences between overall false positive rate and false positive rates for each identity term. FNED calculates the same for false negatives. Results on this dataset are shown in Table \ref{table:synth}. Our method provides substantial improvement on AUC and almost completely eliminates false positive and false negative inequality across identities.

The fine-tuned model also outperforms the baseline for mitigating the bias. The token replacement method comes out as a good baseline for mitigating the bias since it treats all identities the same. The importance weighting approach fails to produce an unbiased model. 

\subsection{Nearest Neighbors of Identity Terms}
Models convert input tokens to embeddings before providing them to convolutional layers. As embeddings make up the majority of the parameters of the network and can be exported for use in other tasks, we're interested in how they change for the identity terms. We show 10 nearest neighbors of the terms \textless{}id\textgreater{} (for the token replacement method), ``gay'', and ``homosexual'' -- top two identity terms with the most mean attribution difference (our method vs. baseline), in Table \ref{table:nn}.

The word embedding of the term ``gay'' shifts from having swear words as its neighbors to having the \textless{}pad\textgreater{} token as the closest neighbor. Although the term ``homosexual'' has lower mean attribution, its neighboring words are still mostly swear words in the baseline embedding space. ``homosexual'' also moved to more neutral terms that shouldn't play a role in deciding if the comment is toxic or not. Although they are not as high quality as one would expect general-purpose word embeddings to be possibly due to data size and the model having a different objective, the results show that our method yields inherently unbiased embeddings. It removes the necessity to initialize word embeddings with pre-debiased embeddings as proposed in \newcite{Bolukbasi:2016}. 

The importance weighting technique penalizes the model on the sentence level instead of focusing on the token level. Therefore, the word embedding of ``gay'' doesn't seem to shift to neutral words. The token replacement method, on the other hand, replaces the identity terms with a token that is surrounded with neutral words in the embedding space, so it results in greater improvement on the synthetic dataset. However, since all identity terms are collapsed into one, it's harder for the model to capture the context and as a result, classification performance on the original dataset drops.

\begin{table}
\resizebox{0.48\textwidth}{!}{%
    \begin{tabular}{c|cc|cc|cc}
     \textbf{Ratio} & \multicolumn{2}{c|}{\textbf{$1\%$}} & \multicolumn{2}{c|}{\textbf{$5\%$}} & \multicolumn{2}{c}{\textbf{$10\%$}} \\ \hline
    \textbf{Toxic} & \textbf{Base}  & \textbf{Ours} & \textbf{Base} & \textbf{Ours} & \textbf{Base} & \textbf{Ours} \\ \hline
    hell            & -.002 &  .035  & .002 & .673  &  .076 & .624           \\ \hline
    moron            & -.002 & .044 & .002 & .462  & .077 & .290              \\ \hline
    sh*t            & -.003 & .078 & .006 & .575 & .098 & .437            \\ \hline
    f*ck            & -.003     & .142 & .013 & .643 & .282  &  .682      \\ \hline
    b*tch            & -.003     & .051 & .002 & .397 &  .065 & .362
    \end{tabular}}
    \caption{Subset of toxic terms we used in the experiments and their mean attribution value on the test set for different training sizes.}
    \label{table:toxic}
\end{table}

\subsection{Results on Incorporating Priors in Different Training Sizes}
\begin{figure}
\begin{tikzpicture}
\begin{axis}[
    xlabel={Training Data Percentage},
    ylabel={Test Accuracy},
    xmin=0, xmax=40,
    ymin=90, ymax=96,
    xtick={0,5,10,15,20,25,30,35,40},
    ytick={90,92,94,96},
    legend pos=south east,
]
 
\addplot[
    color=blue,
    mark=*,
    ]
    coordinates {
    (1,90.4)(5,90.9)(10,92.3)(20,94.2)(30,93.8)(40,94.3)
    };
    \addlegendentry{Baseline}
    \addplot[
    color=red,
    mark=x,
    ]
    coordinates {
    (1,92.5)(5,93.9)(10,93.6)(20,94.2)(30,94.4)(40,94.4)
    };
    \addlegendentry{Our method}
    \addplot [
    domain=0:40, 
    samples=100, 
    color=black,
    mark = none, dashed,
    ]
    {93.9};
    \addlegendentry{Rule based}
 
\end{axis}

\end{tikzpicture}
\caption{Test accuracy for different training sizes. The rule based method gives positive prediction if the comment includes any of the toxic temrs.}
\label{fig:M1}
\end{figure}
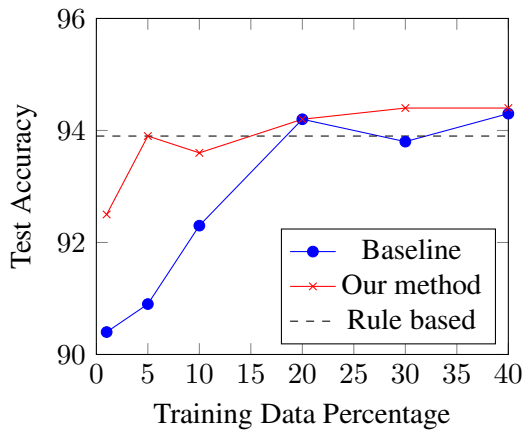

We now demonstrate our approach on encouraging higher attributions on toxic words to increase model performance in scarce data regime. We down-sample the dataset with different ratios to simulate a data scarcity scenario. To directly validate the effectiveness of prior loss on attributions, we first show that the attribution of the toxic words have higher values for our method across different data ratios compared to the baseline in Table \ref{table:toxic}. We also show that the attribution for these terms increases as training data increases for the baseline method. We then show model performance on testing data for different data size ratios for the baseline and our method in Figure \ref{fig:M1}. Our method outperforms the baseline by a big margin in $1\%$ and $5\%$ ratio. However, the impact of our approach diminishes after adding more data, since the model starts to learn to focus on toxic words itself for predicting toxicity without the need for prior injection. We can also see that both the baseline and our method start to catch up with the rule based approach, where we give positive prediction if the toxic word is in the sentence, and eventually outperform it.

\section{Discussion and Related Work}
For explaining ML models, recent research attempts offer techniques ranging from building inherently interpretable models \cite{Kim:2014} to building a proxy model for explaining a more complex model \cite{Ribeiro:2016, Frosst} to explaining inner mechanics of mostly uninterpretable neural networks \cite{pmlr-v70-sundararajan17a, Bach2015OnPE}. One family of interpretability methods uses sensitivity of the network with respect to data points \cite{koh17a} or features \cite{Ribeiro:2016} as a form of explanation. These methods rely on small, local perturbations and check how a network's response changes. Explaining text models has another layer of complexity due to a lock of proper technique to generate counterfactuals in the form of small perturbations. Hence, interpretability methods tailored for text are quite sparse \cite{PramodVQA, JiaLiang, Murdoch}.

On the other hand, there are many papers criticizing the aforementioned methods by questioning their faithfulness, correctness \cite{Adebayo2018SanityCF, Kindermans2017TheO} and usefulness. \newcite{SmilkovTKVW17} show that gradient based methods are susceptible to saturation and can be fooled by adversarial techniques. Other sets of papers \cite{Miller2019ExplanationIA, GilpinBYBSK18} attack model explanation papers from a philosophical perspective. However, the lack of actionability angle is often overlooked. \newcite{Lipton:2018} briefly questions the practical benefit of having model explanations from a practitioners perspective. There are several works taking advantage of model explanations. Namely, using model explanations to aid doctors in diagnosing retinopathy patients \cite{SAYRES2018}, and removing minimal features, called pathologies, from neural networks by tuning the model to have high entropy on pathologies \cite{Feng}. The authors of \newcite{Ross:2017} propose a similar idea to our approach in that they regularize input gradients to alter the decision boundary of the model to make it more consistent with domain knowledge. However, the input gradients technique has been shown to be an inaccurate explanation technique \cite{Adebayo2018SanityCF}.

Addressing and mitigating bias in NLP models are paramount tasks as the effects on these models adversely affect protected subpopulations \cite{Schmidt}. One of the earliest works is \newcite{Calders}. Later, \newcite{Bolukbasi:2016} proposed to unbias word vectors from gender stereotypes. \newcite{Park2018ReducingGB} also try to address gender bias for abusive language detection models by debiasing word vectors, augmenting more data and changing model architecture. While their results seem to show promise for removing gender bias, their method doesn't scale for other identity dimensions such as race and religion. The authors of \newcite{DixonLSTV18} highlight the bias in toxic comment classifier models originating from the dataset. They also supplement the training dataset from Wikipedia articles to shift positive class imbalance for sentences containing identity terms to dataset average. Similarly, their approach alleviates the issue to a certain extent, but does not scale to similar problems as their augmentation technique is too data-specific. Also, both methods trade original task accuracy for fairness, while our method does not. Lastly, there are several works \cite{DavidsonWMW17, Zhang2018DetectingHS} offering methodologies or datasets to evaluate models for unintended bias, but they fail to offer a general framework. 

One of the main reasons our approach improves the model in the original task is that the model is now more robust thanks to the reinforcement provided to the model builder through attributions. From a fairness angle, our technique shares similarities with adversarial training \cite{ZhangBias, MadrasCPZ18} in asking the model to optimize for an additional objective that transitively unbiases the classifier. However, those approaches work to remove protected attributes from the representation layer, which is unstable. Our approach, on the other hand, works with basic human-interpretable units of information -- tokens. Also, those approaches propose to sacrifice main task performance for fairness as well. 

While our method enables model builders to inject priors to aid a model, it has several limitations. In solving the fairness problem in question, it causes the classifier to not focus on the identity terms even for the cases where an identity term itself is being used as an insult. Moreover, our approach requires prior terms to be manually provided, which bears resemblance to blacklist approaches and suffers from the same drawbacks. Lastly, the evaluation methodology that we and previous papers \cite{DixonLSTV18, Park2018ReducingGB} rely on are based on a synthetically-generated dataset, which may contain biases of the individuals creating it. 

\section{Conclusion and Future Work}
In this paper, we proposed actionability on model explanations that enable ML practitioners to enforce priors on their model. We apply this technique to model fairness in toxic comment classification. Our method incorporates Path Integrated Gradients attributions into the objective function with the aim of stopping the classifier from carrying along false positive bias from the data by punishing it when it focuses on identity words.

Our experiments indicate that the models trained jointly with cross-entropy and prior loss do not suffer a performance drop on the original task, while achieving a better performance in fairness metrics on the template-based dataset. Applying model attribution as a fine-tuning step on a trained classifier makes it converge to a more \emph{debiased} classifier in just a few epochs. Additionally, we show that model can be also forced to focus on pre-determined tokens.

There are several avenues we can explore as future research. Our technique can be applied to implement a more robust model by penalizing the attributions falling outside of tokens annotated to be relevant to the predicted class. Another avenue is to incorporate different model attribution strategies such as DeepLRP \cite{Bach2015OnPE} into the objective function. Finally, it would be worthwhile to invest in a technique to extract problematic terms from the model automatically rather than providing prescribed identity or toxic terms.
\section*{Acknowledgments}
We thank Salem Haykal, Ankur Taly, Diego Garcia-Olano, Raz Mathias, and Mukund Sundararajan for their valuable feedback and insightful discussions.

\bibliography{acl2019}
\bibliographystyle{acl_natbib}

\appendix

\end{document}